\documentclass{article}
\usepackage{iclr2025_conference}
\iclrfinalcopy

\setcitestyle{square}
\usepackage{graphicx}
\usepackage{xspace}
\usepackage{hyperref}
\usepackage{listings}
\usepackage{subcaption}

\usepackage{tikz}
\usetikzlibrary{patterns}
\usetikzlibrary{shapes.geometric, arrows}
\usepackage{pgfplots}
\pgfplotsset{compat=1.18}
\usepackage{pgfplotstable}
\pgfplotstableset{
    format=inline,
    row sep=\\,
}

\usepackage{booktabs}
\usepackage{threeparttable}
\usepackage{multirow}

\newif\ifarxiv
\arxivtrue

\usepackage[Tol,OkabeIto]{colorblind}
\usepackage{pifont}

\usepackage{todonotes}
\presetkeys%
    {todonotes}%
    {inline}{}

\title{Guidelines for Applying RL and MARL\\ in Cybersecurity Applications}
\usepackage{authblk}
\author[1]{Vasilios Mavroudis}
\author[2]{Gregory Palmer}
\author[3]{Sara Farmer}
\author[4]{Kez Smithson Whitehead}
\author[5]{David Foster}
\author[6]{Adam Price}
\author[7]{Ian Miles}
\author[1]{Alberto Caron}
\author[1]{Stephen Pasteris}

\affil[1]{Alan Turing Institute, \texttt{vmavroudis@turing.ac.uk}, \texttt{acaron@turing.ac.uk}, \texttt{spasteris@turing.ac.uk}}
\affil[2]{BAE Systems, \texttt{gregory.palmer@baesystems.com}}
\affil[3]{Defence Science and Technology Laboratory, \texttt{sjfarmer@dstl.gov.uk}}
\affil[4]{BMT Group, \texttt{kez.smithsonwhitehead@uk.bmt.org}}
\affil[5]{Applied Data Science Partners, \texttt{david@adsp.ai}}
\affil[6]{Montvieux, \texttt{adam.price@montvieux.com}}
\affil[7]{Frazer-Nash Consultancy, \texttt{i.miles@fnc.co.uk}}

\begin{document}

\maketitle

\section{Introduction}
Reinforcement Learning (RL) and Multi-Agent Reinforcement Learning (MARL) offer promising solutions for complex, dynamic environments where decision-making entities must interact and adapt. In cybersecurity, particularly in Automated Cyber Defence (ACD), these techniques can address challenges posed by high-dimensional observations and actions. This document provides guidelines for:
\begin{itemize}
    \item Cybersecurity professionals exploring RL and MARL for real-world applications.
    \item RL and MARL researchers aiming to tackle the nuanced demands of cybersecurity scenarios.
\end{itemize}

By outlining when RL and MARL are appropriate, addressing cyber-specific challenges, and offering practical considerations for implementation, these guidelines aim to bridge the gap between theoretical research and practical deployment in adversarial settings. We expect that this document will offer support to researchers who are keen to explore topics at the intersection of RL, MARL and ACD by highlighting open research questions and topics that demand further investigation.

\section{Definitions}
\noindent\textbf{Reinforcement Learning (RL):} A learning paradigm where an agent learns to make decisions by performing actions in an environment to maximize cumulative rewards \citep{sutton2018reinforcement}. The actions one takes determine (perhaps stochastically) what one learns about the environment. Specifically, at each point in time one is in a particular state and takes a particular action. The combination of the current state and action determines (perhaps stochastically) the current reward and the next state.

\noindent\textbf{Multi-Agent Reinforcement Learning (MARL):} An extension of RL involving multiple agents learning policies in a shared environment, where each agent can make decisions independently based on its observations \citep{hernandez2018multiagent}.

\noindent\textbf{Multi-Agent Environment:} A setting where multiple agents interact, each with its observations, actions, and possibly objectives. Interactions can be cooperative, competitive, or both, and the environment can be dynamic and adversarial. The MARL literature frequently focuses on three types of games that are defined by the nature of the reward function: 
1)~(Fully)-cooperative games, i.e., team-games, where all agents receive the same reward; 
2)~(Purely) Competitive (Zero-Sum) Games, where the reward received by all players adds up to 0, and; 
3)~General-Sum Games, where the payoffs can be arbitrary \citep{nowe2012game}. While the cyber-defence learning challenge is by definition an adversarial game between Blue cyber-defence and Red cyber-attacking agents, in this work we focus on the learning challenges faced by teams of Blue cyber-defence agents. Therefore, we treat the Red adversary as part of the environment, and discuss challenges, solutions, and opportunities within fully-cooperative cyber-defence scenarios.

\section{Determining if ML Is Appropriate}
Before considering RL or MARL, assess whether ML suits your cybersecurity problem:

\noindent\textbf{Rule-Based Solutions vs. ML.} Many tasks in cybersecurity can be effectively addressed using simple, hardcoded rules. If a problem has a clear, deterministic solution based on well-understood expert knowledge, ML is unnecessary. A good rule of thumb is to ask whether the steps to a solution are straightforward to define. If they are, a rule-based approach is likely more appropriate, as it is simpler, more interpretable, and often more efficient.

\noindent\textbf{Selecting the Right Paradigm.} ML excels at solving problems that align with established learning paradigms, such as supervised learning, reinforcement learning, or unsupervised learning. For ML to succeed, the problem must be carefully modeled to fit one of these paradigms. If a problem does not naturally map to an existing ML paradigm—or no method exists to reframe it appropriately—ML may not be a viable solution unless a novel learning methodology is developed to address the specific challenge.

\noindent\textbf{Explainability Requirements.} When explainability is a critical requirement, ML remains a potential solution, but the choice of techniques becomes more limited. Models must either be inherently interpretable or amenable to post-hoc explainability methods. This restricts the range of applicable algorithms, emphasizing simpler or specially designed models that can generate human-understandable decisions~\citep{10136827}.

\section{Determining if RL Is Appropriate}
Once you have established that your problem is suitable for ML, use the following guidelines to evaluate whether reinforcement learning (RL) is the appropriate approach:

\noindent\textbf{Exploration is Required.} RL is suitable when you do not have a complete or accurate model of the environment (e.g., a Markov Decision Process (MDP) or a Partially Observable MDP (POMDP)). In such cases, the agent must navigate the environment, observe the outcomes of its actions, and adapt its strategy based on these interactions. This ability to explore and learn from the environment is a key differentiator of RL.

\noindent\textbf{Planning is Essential.} RL is the right choice when solving the problem requires non-trivial planning in addition to exploration. In cybersecurity, for instance, defensive actions often need to be executed in a complex sequence ($>1$ action) to achieve strategic objectives. Problems requiring only a single action or where actions do not influence future states can often be addressed with simpler approaches, such as multi-armed bandit algorithms, rather than full RL~\cite{bouneffouf2023multi}.

\noindent\textbf{State/Observation.} Space Considerations. The size and structure of the state or observation space significantly influences the choice of RL algorithms:
\begin{itemize}
    \item \textbf{Small, Discrete State Spaces:}
        For problems with a small, finite state space, policies can often be maintained explicitly using algorithms like UCB-VI~\cite{azar2017minimax}, which offer theoretical performance guarantees.
        
    \item \textbf{Large or Infinite State Spaces:}
        When state spaces are large or continuous, function approximators such as neural networks become necessary. 
In these cases: (1) Deep RL (DRL) algorithms like policy gradient methods or deep Q-learning are commonly used~\cite{wang2022deep}; (2) Explicit algorithms designed for discrete spaces typically fail to scale effectively.
\end{itemize}

By considering these factors, you can assess whether RL is the right paradigm to address your specific problem and determine the appropriate algorithmic approach.
In addition, depending on the target scenario, DRL agents may also be confronted with the challenge of the size of the state/observation space not being known ā priōrī. For example, the number of nodes on a target network may not be known in advance, or may even fluctuate during deployment. This raises the need for function approximators that can handle variable sized inputs \citep{symes2023entity,palmer2023deep}.

\subsection{Additional Considerations}

In addition to the above, the following considerations should be made before selecting an approach:
\begin{itemize}
    \item \textbf{Markov Decision Process (MDP):} Can your problem be modelled as an MDP? (Where the next state depends only on the current state and action independent of the past.)
    
    \item \textbf{Partially Observable MDP (POMDP):} If the agent cannot fully observe the state, but instead only observes partial information that may help it determine the state, then memory (of the previous observations) will help it infer the current state. Hence, consider whether memory mechanisms are required.

    \item \textbf{Exploration vs. Planning:} Determine whether your problem requires the agent to explore unknown states or mainly focuses on planning in a known environment.
\end{itemize}

\section{Determining whether MARL is Necessary}
Multi-Agent Reinforcement Learning (MARL) offers powerful solutions for complex environments but is challenging to implement and resource-intensive. Before committing to a MARL-based solution, carefully evaluate whether multiple agents are truly necessary. Simpler alternatives, such as single-agent RL or hierarchical structures, may suffice for certain problems. MARL is particularly effective in scenarios with:

\begin{itemize}
    \item Large and Complex Observation/Action Spaces: Distributed agents can manage intricate state and action spaces that a single agent would struggle to handle.

    \item Fragmented Networks: In environments where networks are divided due to link failures or physical separation, MARL enables localized decision-making.

    \item Resilience Requirements: Distributed agents ensure continuity in critical applications, such as military or infrastructure defence, even if some fail.

    \item Scalability Needs: MARL supports adding agents dynamically without overhauling the system.

    \item Weak Dependencies Between Sub-Networks: Agents can operate semi-independently in loosely connected sub-networks, reducing the complexity of coordination.
\end{itemize}

\subsection{When MARL May Be Excessive}
Not all problems justify MARL’s complexity. If MARL is being considered solely to accelerate single-agent methods like Proximal Policy Optimization (PPO), it is unlikely to deliver significant benefits. Similarly, tasks with simple structures or clear solutions may not require multiple agents.

\subsubsection{Understanding the Environment}
The structure and dynamics of the environment determine whether MARL is appropriate:
\begin{itemize}
    \item Environment Types: When each agent has full information about the current state, multi-agent environments often manifest as Decentralised MDPs (Dec-MDPs), where there is a single shared reward and hence agents fully cooperate tasks~\citep{ oliehoek2016concise}, or Markov Games, where each agent has its own reward to maximise (implying a mixed cooperative-competitive setting). However, in reality agents often do not have full information about the current state, instead observing some form of information that aids them in deducing it. When given this partial-observability, the above two models are known as “Decentralised partially observable MDPs” (Dec-POMDPs) and “Partially observable Markov games” (POMGs) respectively.

    \item Cybersecurity Context: Blue vs. Red scenarios frequently fall under POMGs, where adversaries (Red) are treated as part of the environment, simplifying analysis from the Blue team’s perspective.
\end{itemize}

\subsection{Communication and Coordination}
Effective communication between agents is often critical in MARL:

\begin{itemize}
    \item \textbf{Communication Needs:} Does your problem require agents to exchange information? If so, consider limitations such as bandwidth, latency, or the risk of communication link failures.
    \item \textbf{Coordination Complexity:} Environments requiring high levels of inter-agent coordination can quadratically increase data and computational requirements.
\end{itemize}

\subsection{Resource Considerations}
MARL systems demand substantial resources, both in terms of data and computational power:
\begin{itemize}
    \item \textbf{Data Requirements:} MARL’s data needs increase with the number of agents and the complexity of coordination. Cooperative tasks exacerbate these demands.

    \item  \textbf{Reward Design:} Properly defining reward functions and performance metrics is essential to guide agent behavior effectively.

    \item \textbf{Timeliness:} Evaluate if communication delays or processing time could hinder real-time responses, such as containing a cyber-attack.
\end{itemize}

\section{Challenges in MARL for Cybersecurity Applications}
In cyber-defence scenarios modelled as Partially Observable Markov Games (POMGs) or Decentralized Partially Observable Markov Decision Processes (Dec-POMDPs), Multi-Agent Reinforcement Learning (MARL) becomes a viable approach. However, naive MARL methods often converge on sub-optimal joint policies due to inherent learning pathologies. 
This challenge raise the question of what constitutes a desirable outcome in a MARL context.

From an equilibrium perspective, for fully-cooperative learning problems, an ideal joint policy is Pareto optimal, where no agent can improve its outcome without making another agent worse off. In non-cooperative scenarios, a Nash equilibrium may also be considered, where no agent can unilaterally deviate to a better policy while the others’ policies remain unchanged. The following pathologies commonly hinder convergence to optimal policies in MARL systems:

\noindent\textbf{Relative Overgeneralization.}
Agents may converge to sub-optimal Nash equilibria due to random exploration, which leads to regions in the reward space that are locally optimal but globally sub-optimal \citep{ JMLR:v17:15-417}. This pathology is exacerbated when multiple agents explore independently.

\noindent\textbf{Stochasticity.}
Variability in rewards and state transitions can skew learning processes, drawing agents toward sub-optimal equilibria \citep{JMLR:v17:15-417,palmer2018negative}. This is influenced by the type of MARL algorithm used and the stochastic nature of the environment. For example, within a cyber-defence context a cyber-attacking agent using a stochastic policy may introduce stochasticity. In addition, in order to confront our learners with a wide range opponents, the sampling of available opponents may in itself be a stochastic process. 

\noindent\textbf{Alter-Exploration Problem.}
The probability of at least one agent taking an exploratory action increases with the number of agents. For example, under epsilon-greedy exploration, the likelihood of exploration at a time step is $1-(1-\epsilon)^n$ where n is the number of agents \citep{matignon2012independent}. This often leads to inconsistent joint behaviour during exploration.

\noindent\textbf{Miscoordination.}
Agents may independently select actions that, while optimal for their individual objectives, are incompatible with the collective goal \citep{matignon2007hysteretic}. For instance, in cyber-defence, multiple agents updating a firewall’s access control list simultaneously might result in contradictory or vulnerable configurations.

\noindent\textbf{Moving Target Problem.}
An environment can no longer be considered Markovian when it is inhabited by multiple independent learners who are updating their policies in parallel. The resulting non-stationarity of policies can make an environment dynamic and harder to learn from \citep{ busoniu2008comprehensive,bowling2002multiagent}. 

\subsection{Why Learning in MARL is Hard}
Overcoming these pathologies requires effective exploration strategies. However, exploration inherently relies on random action selection, giving rise to the miscoordination and stochasticity pathologies. Even with well-optimized agents, the likelihood of selecting the optimal sequence of actions during exploration remains low, due to the alter-exploration problem. In addition to the above challenges, achieving a desirable outcome in cyber-defence scenarios can be difficult due to real-world constraints such as limited data availability and computational resources.

By understanding and addressing these pathologies, future MARL methods can better align with the demands of cyber-defence, leading to more robust and effective multi-agent policies. To date the field of ACD has often benefited from custom DRL approaches that incorporate domain knowledge from the target cyber-defence scenarios. Therefore, we hypothesize that identifying particularly challenging cyber-defence scenarios where agents are being confronted by the above pathologies could lead to the development of novel custom built MADRL solutions. In the next section, Algorithmic Considerations, we shall consider a number of MADRL approaches that could serve as a starting point for such developments. 

\subsection{Real-World Cybersecurity Challenges}
There are many fundamental real-world challenges associated with the cybersecurity problem space that need to be considered when applying MARL for this use case. Below is a non-exhaustive list of our research focus areas: 

\noindent\textbf{High-Dimensional Observations:} Cyber environments often have vast and heterogeneous data (e.g., log files).

\noindent\textbf{Partial Observability:} Agents may have a limited view of the network.

\noindent\textbf{Time Constraints:} Actions may have delays, and decisions need to be timely.

\noindent\textbf{Communication Limitations:} Network fragmentation and potential interception by adversaries complicate inter-agent communication.

\noindent\textbf{Integration:} Retrofitting ACD technologies is a challenge as actuation \& integration touchpoints may not be available (e.g. network taps to provide observation data).

\noindent\textbf{Data Limitations:} Access to sufficient and relevant data for training can be challenging. ACD agents are dependent on effective attack detection sources, highlighting the attacker/defender arms race that is a constant challenge in the cybersecurity domain. In addition false positive alerts, user (green agent) traffic and routine equipment failures will introduce noise into agent observations.

Palmer et al. \citep{palmer2023deep} explore characterisation of cyber-defence as an RL problem in more detail.

As technologies mature and we consider exploitation routes there are many other real-world challenges under active research, many of which are common across AI technologies in other domains:

\noindent\textbf{AI Assurance \& Trust} Several studies can be found 
on this topic\footnote{\href{https://www.qinetiq.com/en/capabilities/ai-analytics-and-advanced-computing/autonomous-resilient-cyber-defence}{https://www.qinetiq.com/en/capabilities/ai-analytics-and-advanced-computing/autonomous-resilient-cyber-defence}}, 
while the UK Defence policy is also progressing for the general adoption of AI, captured in Joint Service Publication 936~\citep{mod2024jsp936}.

\noindent\textbf{Human Machine Teaming.} Human science considerations for how agents and humans can work together and be trusted, including consideration of the level of automation. 
For example agents as ‘recommender’ support systems are easier to adopt than fully autonomous capability. 
The availability of human cybersecurity expertise is both a challenge and an opportunity, ACD provides an opportunity for defence of devices that currently have no local human cybersecurity incident responders.

\noindent\textbf{Safety \& Security.} ACD agents will have to operate within the existing constraints of safety and security policy, which will be a particular challenge in application to safety-critical Industrial Control Systems. One approach that will help with this area is the introduction of Authorised Bounds, discussed in this briefing paper\footnote{\href{https://cetas.turing.ac.uk/publications/autonomous-cyber-defence-autonomous-agents}{https://cetas.turing.ac.uk/publications/autonomous-cyber-defence-autonomous-agents}}, 
where a clear policy of permitted agent actions should help limit safety impacts.

\noindent\textbf{Security of AI.} Introduction of ACD agents also introduces a new attack surface to target systems. Robust test and evaluation of ACD agent performance will be required prior to use in operations. Security of AI is a high-profile and rapidly evolving field, which will be monitored to inform future research and assurance. For example, deterministic policies can be vulnerable to adversarial probing; consider rotating policies or introducing randomness. 

\noindent\textbf{Continuous learning.} Decide whether agents will continue learning online, receive periodic updates, or use fixed policies. Online learning will greatly increase assurance, trust and security challenges.

\section{Algorithmic Considerations}

Applying MADRL to cybersecurity requires careful selection of algorithms that align with the specific needs of the environment and the level of coordination among agents. This section explores various algorithmic approaches, their advantages, challenges, and relevance to cyber defence scenarios.

\subsection{Joint Action Learners}
Joint Action Learners (JAL) centralize the learning process by considering the joint action space of all agents. This approach effectively transforms the multi-agent learning problem into a single-agent learning problem with a combined action space, directly addressing coordination by learning a joint policy.
Centralized approaches can find globally optimal solutions by accounting for all agents' actions simultaneously. They handle coordination issues effectively and eliminate problems like miscoordination or conflicting actions. However, this approach can become computationally infeasible as the number of agents increases due to the exponential growth of the joint action space. The sample complexity also increases significantly, requiring more data to explore and learn optimal policies.
In scenarios with a small number of agents, JAL can be effective. For instance, in certain cyber-defence applications where only a few critical nodes need to be coordinated tightly, this approach might be feasible. However, they may not scale well for larger systems common in complex network defence, making them less practical for widespread cybersecurity applications.

\subsection{Parameter Sharing}
A consideration when using MARL, regardless of the algorithm type, is whether agents should share the same policy weights. When agents in a system are identical, i.e. matching observations and action spaces, agents can share the same policy. Naturally, parameter sharing causes matching agents to behave identically, which could be a disadvantage in problems where complex coordination is required. However, if inter-agent interactions are simple, parameter sharing can decrease training time and increase training stability as training batch sizes are effectively multiplied by the number of matching agents. 
Parameter sharing can also be modified to give agents a sense of independence by encoding each agent's identity in the observation space (i.e. each agent’s observation is now composed of its original observation and a unique identifier value corresponding to the agent). Augmenting the observation in this manner can help resolve coordination problems and can serve as a strong benchmark before applying different methods.

\subsection{Independent Learning Methods}
Independent learning methods treat each agent as an autonomous learner within a shared environment. Each agent learns its own policy independently, without direct coordination or communication with other agents. This approach is straightforward to implement and can be effective when minimal cooperation is required among agents.

One example is Independent Proximal Policy Optimization (IPPO) \citep{ yu2022surprising}. In this method, each agent employs Proximal Policy Optimization (PPO) independently while interacting within a Markov game. IPPO is suitable when agents can act independently without significant coordination. Parameter sharing among agents can be considered to reduce computational overhead and promote consistency.

Another category within independent learning methods is the Optimistic Approaches, such as Leniency and Hysteretic Q-learning. These methods address coordination issues through optimism in value estimates:
\begin{itemize}
    \item Leniency allows agents to be forgiving towards negative outcomes during exploration, encouraging continued exploration of potentially beneficial joint actions. By being lenient, agents can avoid prematurely converging to suboptimal policies due to temporary negative feedback \citep{JMLR:v17:15-417,palmer2018lenient}.

    \item Hysteretic Q-learning uses different learning rates for positive and negative rewards. This approach stabilizes learning by making agents more receptive to positive experiences while being less affected by negative ones \citep{ matignon2007hysteretic,lu2018improving,omidshafiei2017deep}.
\end{itemize}

Additionally, Difference Rewards Learning tackles the credit assignment problem by estimating the difference between the global reward and the reward without a specific agent's contribution \citep{foerster2018counterfactual}. This helps determine each agent's individual impact on overall performance. However, in cyber-defence scenarios, it can be challenging to establish the counterfactual without re-running episodes, making it computationally intensive. An alternative is to use Shapley values from cooperative game theory to fairly distribute the total gains among agents. Shapley values have been applied in explainable AI (XAI) and are relevant in assessing agent contributions in multi-agent settings. For instance, projects like ARCD have utilized Shapley values to evaluate the impact of trained agents in cybersecurity simulations.

\subsection{Centralized Training with Decentralized Execution}
CTDE methods involve agents sharing information during training but operating independently during execution. This approach enhances coordination during learning without incurring communication overhead during deployment.

An example is Multi-Agent Proximal Policy Optimization (MAPPO) \citep{ yu2022surprising}, which extends PPO to multi-agent settings by incorporating a centralized critic during training. The centralized critic is conditioned on either global state information or shared observations, helping to coordinate agents by providing shared value estimates. Sharing information through the critic enhances coordination and can lead to improved performance.
However, MAPPO may have issues with agents overfitting to the centralized critic, reducing generalization to unseen environments. Agents might rely too heavily on the critic's guidance during training, which may not be available during execution. A simple trick to prevent overfitting is to add noise to the input to the critic, preventing it from outputting consistent values in early training~\citep{hu2021policy}. This reasonably simple code trick can give a noticeable increase in performance.

Although MAPPO has strong empirical performance, it lacks mechanisms that directly address agent coordination or credit assignment issues. Value Function Factorization (VFF) methods actively try to resolve coordination or credit assignment issues by estimating a joint value function. Similar to JAL, which produces value estimates based on the joint action-observation, value factorisation methods produce the joint Q-function (Qtot) from a set of utility functions. Each agent uses a utility function to estimate its local expectation, given its observations and actions. Different VFF methods combine the outputs of the utility functions to form the joint expectation function in different ways:

Value-Decomposition Networks (VDN)\citep{duarte2018variable} simply sum the individual Q-values of all agents to form the joint Q-value.

QMIX\citep{rashid2018qmix} uses a mixing network to combine individual Q-values into the joint Q-value while ensuring that the joint action-value function is monotonic in each agent's utility function. This allows for the learning of complex joint action-values while still enabling decentralized execution.

These methods help with credit assignment by attempting to calculate each agent's effect on the joint Q-function. However, methods like QMIX rely on the Individual-Global-Max (IGM) assumption, which presumes that when each agent uses greedy action selection, the joint action will also be optimal. This assumption may not hold in environments requiring complex coordination. While value-based methods like QMIX are generally more data-efficient—which can be beneficial in cyber settings—they can be inconsistent in implementation and may not effectively handle coordination in all cases. Studies have pointed out issues with the IGM assumption and suggested that these methods might not scale well in highly cooperative tasks (e.g., Son et al., 2019). In addition, hyperparameter tuning for QMIX can be time-consuming \citep{tran2022cascaded}.

\section{Communication Learning}
Communication learning involves agents learning what and how to communicate to improve coordination. In complex tasks requiring cooperation, communication can be a crucial factor in achieving optimal performance.

Some algorithms enable agents to learn communication protocols during training:

Reinforced Inter-Agent Learning (RIAL) \citep{foerster2016learning} uses recurrent neural networks to allow agents to maintain internal states and learn communication policies. Agents can develop their own communication strategies that are tailored to the task.

Differentiable Inter-Agent Learning (DIAL) \citep{foerster2016learning} facilitates end-to-end learning of communication protocols by making communication channels differentiable. This enables backpropagation through communication actions, allowing agents to optimize their communication along with their policy learning.

The emergence of compositional language is another aspect of communication learning. Agents develop grounded, structured communication resembling human language, enabling them to express complex concepts and adapt to new tasks. Works like Foerster et al.~\citep{foerster2016learning} and Mordatch and Abbeel \citep{mordatch2018emergence} have explored how agents can develop such communication strategies, which can enhance coordination and generalization.

Learned communication protocols introduce additional complexity in training and may pose challenges in cybersecurity contexts:

\begin{itemize}
    \item Training Complexity: Agents must learn both task policies and communication strategies, increasing the learning burden and computational requirements.

    \item Stability Issues: Communication can introduce non-stationarity and convergence problems, as agents' policies and communication protocols evolve simultaneously.

    \item Security Risks: In cybersecurity, communication channels may be susceptible to interception or manipulation by adversaries. Agents need to consider the possibility of eavesdropping or spoofing attacks, which complicates the design of communication protocols.

\end{itemize}

Alternatively, predefined communication protocols involve establishing communication rules before training. This simplifies implementation and reduces training complexity. Agents operate under known communication constraints, which can be advantageous in environments where security and reliability are paramount. However, predefined protocols may limit the flexibility of agents to adapt communication to specific scenarios or to develop more efficient strategies.

\subsection{Model-Based MARL}
Model-based methods aim to improve sample efficiency by learning a model of the environment's dynamics, which agents use for planning or generating synthetic experiences. This includes opponent modelling, where an agent attempts to learn other agents’ strategies and predict their actions. In cyber-defence scenarios, where data may be scarce or expensive to obtain, model-based MARL can potentially offer significant advantages.

In Model-Based Reinforcement Learning (MBRL), agents learn a world model that can predict future states and rewards based on current states and actions. This allows agents to simulate interactions with the environment internally, reducing the need for extensive real-world data. Dreamer and Model-Based Policy Optimization (MBPO) are examples of algorithms that leverage world models for policy learning, in two different ways:

\begin{itemize}
    \item Dreamer relies entirely on imagined trajectories generated by the world model. It allows agents to plan ahead by simulating possible futures and selecting actions that maximize cumulative imaginary returns.

    \item MBPO augments data collected from the environment with data generated from the world model. It combines model-based and model-free learning to reduce model bias and improve performance.
\end{itemize}

Applying model-based methods in MARL presents several challenges:

\begin{itemize}
    \item Model Bias and Compounding Errors: Inaccuracies in the learned model can lead to errors that accumulate over time, resulting in suboptimal policies. This issue is exacerbated in multi-agent settings due to the complexity of modelling interactions among agents.

    \item \textbf{Scalability Issues:} Learning an accurate model of an environment with many interacting agents is computationally intensive. The state and action spaces grow exponentially with the number of agents, making it difficult to capture all relevant dynamics.

    \item \textbf{Decentralization Assumptions: }In MARL, agents' actions can significantly affect each other's states. If the assumption of independent dynamics does not hold, decentralized models may fail to capture critical interactions, leading to poor performance.

    \item \textbf{Data Requirements:} Accurate models of environment dynamics can require a large amount of training data, an issue that could be compounded in a multi-agent case.  
\end{itemize}

Despite these challenges, model-based MARL remains an open area of research with potential benefits:

\begin{itemize}
    \item \textbf{Sample Efficiency:} By leveraging simulated experiences, agents can learn effective policies with fewer interactions with the real environment.

    \item \textbf{Sim-to-Real Transfer: }Incorporating real data into world models can help bridge the gap between simulation and real-world deployment, enhancing the applicability of trained agents in actual cyber-defence scenarios.
    
    \item \textbf{Adaptive Planning:} Agents can quickly adapt to changes in the environment by updating the world model, making them more resilient to evolving threats.
\end{itemize}

\section{Lessons Learned}

\noindent\textbf{Accurate Reward Functions.} Ensure rewards align with real-world objectives to produce applicable policies.

\textit{Common pitfall} - Creating rewards that encourage ‘gaming’ the system rather than achieving true security goals.\\
\textit{Solution} - Work closely with security experts to define accurate reward structures that reflect reality.\\

\noindent\textbf{Sparse Rewards.} Cyber environments often provide sparse feedback; designing informative rewards is crucial.

\textit{Common pitfall }- Agents may go long periods without receiving meaningful feedback.\\
\textit{Solution} - Implement reward shaping to provide intermediate feedback.\\

\noindent\textbf{Credit Assignment.} Identifying each agent's contribution helps in refining policies and improving cooperation.

\textit{Common pitfall} - Agents may have overlapping responsibilities.\\
\textit{Solution} - Design local reward functions that complement but do not overshadow global objectives.\\

\noindent\textbf{Generalization and Robustness.} Agents should be trained to handle variability in the environment and adversary behaviors.

\textit{Common pitfall} - Agents do not generalise to a range of scenarios.\\
\textit{Solution} - Ensure the environment is highly configurable and train agents on a diverse array of environment parameters, not just a narrow set of scenarios.\\

\noindent\textbf{Interdisciplinary Insights.} Leverage findings from game theory and autonomy research to anticipate and mitigate common issues.

\textit{Common pitfall} - RL may not sufficiently explore the policy space using traditional methods.\\
\textit{Solution} - Leverage learnings from game theory and other fields to develop methods that allow agents to exploit novel strategies.\\

\noindent\textbf{Real-world challenges.} ACD research can struggle to bridge the ‘Sim to Real’ gap and become beneficial in the real world.

\textit{Common pitfall}: Assumed dependencies (such as detection alerts), and actuation points may be noisy, unreliable or unavailable.\\
\textit{Solution}: Work closely with domain experts to validate and refine assumptions throughout the research, testing in increasingly high-fidelity environments.\\

\section{Next Steps and Open Research Questions}

\noindent\textbf{Policy Optimality:} Investigate whether current MARL methods can achieve Pareto-optimal policies in cyber-defence scenarios.

\noindent\textbf{Cooperation Levels:} Determine the necessary level of cooperation among agents for effective defence.

\noindent\textbf{Contribution Analysis:} Use methods like Shapley values to assess individual agent contributions.

\noindent\textbf{Scalability:} Explore how increasing the number of agents affects learning and performance.

\noindent\textbf{Communication Protocols:} Evaluate the effectiveness of learned versus predefined communication strategies.

\noindent\textbf{Heterogeneous Agents:} Assess the impact of agents with different roles and action spaces on learning outcomes.

\noindent\textbf{Generalization Across Adversaries:} Test the robustness of trained policies against various types of attackers.

\noindent\textbf{Order of Actions:} Examine how the sequence of agent actions influences overall effectiveness.

\noindent\textbf{Real-World Integration:} Continue testing and evaluating agent performance in increasingly representative environments.

\section{Conclusion}
Applying RL and MARL in cybersecurity presents both opportunities and challenges. While much of this document has focused on fully cooperative team game scenarios from the defender's (Blue team) perspective, it's important to acknowledge that this assumption doesn't always hold in real-world settings. For instance, different third parties may manage various parts of a network, leading to self-interested agents that prioritize protecting their own resources. This shifts the problem from cooperative team games to general-sum games, potentially giving rise to sequential social dilemmas. In such cases, converging on a cooperative joint policy may not align with each agent's best interests, and individuals might choose actions that benefit themselves at the expense of collective goals, such as hoarding resources.
Recognizing these complexities underscores the importance of carefully assessing the suitability of ML approaches and understanding the nuances of MARL in cyber environments. By considering scenarios in which agents may not fully cooperate and addressing the unique challenges these situations present, practitioners can develop more robust and effective solutions for automated cyber defence. Collaboration between cybersecurity professionals and MARL researchers is essential to advance the state of the art, tackle open research questions, and create systems capable of handling the diverse and dynamic nature of real-world networks.

\section*{Ethical considerations}
Ethical considerations and adherence to security policies are crucial when implementing autonomous systems in cybersecurity. Always ensure compliance with relevant guidelines and best practices. The latest defence guidance on ethical challenges associated with AI/autonomy is available in the recently released JSP 936~\citep{mod2024jsp936}. 

\section*{Acknowledgements} 
This paper represents research funded by the Defence Science and Technology Laboratory (DSTL), which is an executive agency of the UK Ministry of Defence providing world class expertise and delivering cutting-edge science and technology for the benefit of the nation and allies. The  research supports the Autonomous Resilient Cyber Defence (ARCD) project within the Dstl Cyber Defence Enhancement programme. We would like to thank Isaac Thompson for his contributions in the initial stages of this work.

\bibliographystyle{plainnat}
\bibliography{references}

\newpage
\appendix

\section{Implementing RL and MARL}
Selecting the right framework is critical for success. It can influence scalability, ease of use, documentation quality, and overall development efficiency. This section provides a comparative overview of popular RL and MARL frameworks, highlighting their key advantages, drawbacks, and trade-offs to help practitioners and researchers make informed decisions. Table~\ref{tab:framework_comparison} summarizes these frameworks, offering insights into their capabilities and limitations for tackling complex, real-world applications such as automated cyber defence.

\begin{table}[h!]
    \centering
\resizebox{\columnwidth}{!}{
    \begin{tabular}{|l|p{6cm}|p{6cm}|}
        \hline
        \textbf{Framework} & \textbf{Pros} & \textbf{Cons} \\ \hline
        \href{https://docs.ray.io/en/latest/rllib/index.html}{RLlib} & Compute scaling (e.g., cluster compute), customisable. & Learning Curve, poor docs, often updating, large overhead. \\ \hline
        \href{https://stable-baselines3.readthedocs.io/en/master/}{SB3} & Robust and thoroughly tested implementations. Widely adopted and easy to use. & Not multi-agent out of the box. Not very well maintained, often lacks optional features (e.g., curiosity-ICM). \\ \hline
        \href{https://github.com/vwxyzjn/cleanrl}{CleanRL} & Lightweight - Single file implementation of RL algorithms. Multi-agent out of the box. Adoption is increasing. & Not production grade yet, ok for research. \\ \hline
        \href{https://www.agilerl.com/}{AgileRL} & Uses evolutionary hyperparameter optimization and allows for distributed training. & Not widely adopted. \\ \hline
        \href{https://github.com/instadeepai/Mava}{Mava} and \href{https://arxiv.org/abs/2311.10090}{JaxMARL} & Fast training. & Sharp edges of JAX. \\ \hline
        \href{https://spinningup.openai.com/en/latest/}{SpinningUp} & Clean, lightweight implementation and great accompanying documentation. Designed to educate the user. & Only Single Agent RL. No Multi-Agent. \\ \hline
        DIY & Tailored implementation, deeper understanding, full control. & Max pain. Less efficient, reinventing the wheel. \\ \hline
    \end{tabular}}
    \caption{Comparison of MARL Frameworks}
    \label{tab:framework_comparison}
\end{table}

\section{RL Interfaces}
RL interfaces provide a consistent way to interact with environments, making it easier to implement and test RL algorithms across different domains. In the context of cybersecurity and automated cyber defence, choosing the appropriate interface can significantly influence scalability, flexibility, and performance. This section introduces prominent RL interfaces, highlighting their functionality, features, and repositories for further exploration. Table~\ref{tab:rl_interfaces} summarizes these tools, offering guidance on selecting interfaces that align with single-agent and multi-agent use cases, including centralized and scalable MARL systems.

\begin{table}[h!]
    \centering
    \begin{tabular}{|l|p{10cm}|}
        \hline
        \textbf{Name} & \textbf{Description} \\ \hline
        Gymnasium \citep{Towers_Gymnasium_A_Standard} & Standard interface for single-agent reinforcement learning. It can be used for centralised MARL. \\ \hline
        PettingZoo \citep{Terry_PettingZoo_Gym_for} & Multi-agent native version of Gymnasium.  \\ \hline
        Entity Gym \citep{entity_gym} & Extends the standard Gym interface to allow potentially variable-length lists of ‘entities’ or objects (read-nodes) in observation and action space. It allows for a form of centralised MARL that scales well with number of entities. \\ \hline
    \end{tabular}
    \caption{RL Interfaces and their Repositories}
    \label{tab:rl_interfaces}
\end{table}

\section{Cyber Environments for RL \& MARL}
Environments allow researchers and practitioners to test RL agents in controlled, explainable scenarios that mimic real-world adversarial challenges. In Automated Cyber Defence, such environments simulate network architectures, security threats, and defensive mechanisms, enabling the design and assessment of scalable RL and MARL solutions. This section highlights several key cyber environments that have been widely used in ACD research. The following environments are common in ACD research and testing:

\begin{itemize}
\item \href{https://github.com/dstl/YAWNING-TITAN}{YAWNING TITAN}: An abstract, graph-based cyber-security simulation environment.
\item \href{https://github.com/alan-turing-institute/CybORG_plus_plus/}{CAGE2 CyBORG}: An open-source single-agent cyber challenge environment.
\item \href{https://github.com/cage-challenge/cage-challenge-4}{CAGE4 CyBORG}: An open-source multi-agent cyber challenge environment.
\item \href{https://arxiv.org/abs/2409.10563}{IPMSRL}: A maritime platform management system simulator. Currently this environment is not open source but its wider availability may improve in 2025. 
\item \href{https://generaldynamics.uk.com/systems/mission-systems/land-products/generic-vehicle-architecture/}{Generic Vehicle Architecture (GVA)}: Simulates vehicle network architectures with message passing. Currently this environment is not openly available and is heavily integrated with its owner's infrastructure.
\item PrimAITE V3 expands PrimAITE V2 to enable MARL but is not yet open source. This may change in 2025. 
\end{itemize}

It is also noted that the \href{https://github.com/Limmen/awesome-rl-for-cybersecurity}{awesome-rl-for-cybersecurity} repo is a great resource for RL environments and more.

\end{document}